\pdfoutput=1

\documentclass[11pt]{article}

\usepackage[final]{acl}

\usepackage{times}
\usepackage{latexsym}

\usepackage[T1]{fontenc}

\usepackage[utf8]{inputenc}

\usepackage{microtype}

\usepackage{inconsolata}

\usepackage{graphicx}
\usepackage{tabularx}
\usepackage{makecell}
\usepackage{booktabs}
\usepackage{algorithm}
\usepackage{algpseudocode}
\usepackage{soul,color,xcolor}
\usepackage{longtable}
\usepackage{arydshln}
\usepackage{array}
\usepackage{enumitem}

\definecolor{myyellow}{rgb}{1.0,1.0,0.6}
\definecolor{blue}{RGB}{187,227,248}
\definecolor{purple}{RGB}{212,187,248}
\newcommand{\hlcustom}[2]{\sethlcolor{#1}\hl{#2}\sethlcolor{myyellow}}
\newcommand{\hlblue}[1]{\hlcustom{blue}{#1}}
\newcommand{\hlpurple}[1]{\hlcustom{purple}{#1}}

\newcommand\blfootnote[1]{
  \begingroup
  \renewcommand\thefootnote{}\footnote{#1}
  \addtocounter{footnote}{-1}
  \endgroup
}

\title{Towards Enhanced Immersion and Agency for LLM-based\\Interactive Drama}

\author{Hongqiu Wu\textsuperscript{\rm 1,3,4}, Weiqi Wu\textsuperscript{\rm 1,3,4}, Tianyang Xu\textsuperscript{\rm 1,3,4}, Jiameng Zhang\textsuperscript{\rm 2}, Hai Zhao$^\dag$\textsuperscript{\rm 1,3,4} \\
  \textsuperscript{\rm 1}School of Computer Science, Shanghai Jiao Tong University
  \textsuperscript{\rm 2}University of Zurich \\
  \textsuperscript{\rm 3} Key Laboratory of Shanghai Education Commission for Intelligent Interaction\\ and Cognitive Engineering, Shanghai Jiao Tong University \\
  \textsuperscript{\rm 4} Shanghai Key Laboratory of Trusted Data Circulation and Governance in Web3 \\
  \texttt{\{wuhongqiu,wuwq1022,johnnie.walker\}@sjtu.edu.cn,} \\
  \texttt{jiameng.zhang@uzh.ch,zhaohai@cs.sjtu.edu.cn} \\
  }

\begin{document}
\maketitle

\blfootnote{$^\dag$ Corresponding author. This research was supported by the Joint Research Project of Yangtze River Delta Science and Technology Innovation Community (No. 2022CSJGG1400). Code and demonstration are in \url{https://github.com/gingasan/interactive-drama}.}

\begin{abstract}
\emph{LLM-based Interactive Drama} is a novel AI-based dialogue scenario, where the user (i.e. the player) plays the role of a character in the story, has conversations with characters played by LLM agents, and experiences an unfolding story. This paper begins with understanding interactive drama from two aspects: \textbf{Immersion}—the player's feeling of being present in the story—and \textbf{Agency}—the player's ability to influence the story world. Both are crucial to creating an enjoyable interactive experience, while they have been underexplored in previous work.
To enhance these two aspects, we first propose \textit{Playwriting-guided Generation}, a novel method that helps LLMs craft dramatic stories with substantially improved structures and narrative quality. Additionally, we introduce \textit{Plot-based Reflection} for LLM agents to refine their reactions to align with the player's intentions. Our evaluation relies on human judgment to assess the gains of our methods in terms of immersion and agency.
\end{abstract}

\section{Introduction}

Human existence is deeply rooted in emotional expression and connections. AI-based dialogue systems can provide instant and cheap companionship, where humans are free to express themselves. However, such open-ended interactions fall short of building a deep and complete emotional connection, which naturally emerges from shared experiences. This gap can be bridged by co-constructing an enjoyable interactive experience between humans and AI. This paper studies \textit{LLM-based Interactive Drama}, a new format of dialogue application, which allows the human player to experience a story with dramatic development.

\begin{figure}[t]
\centering
\includegraphics[width=0.49\textwidth]{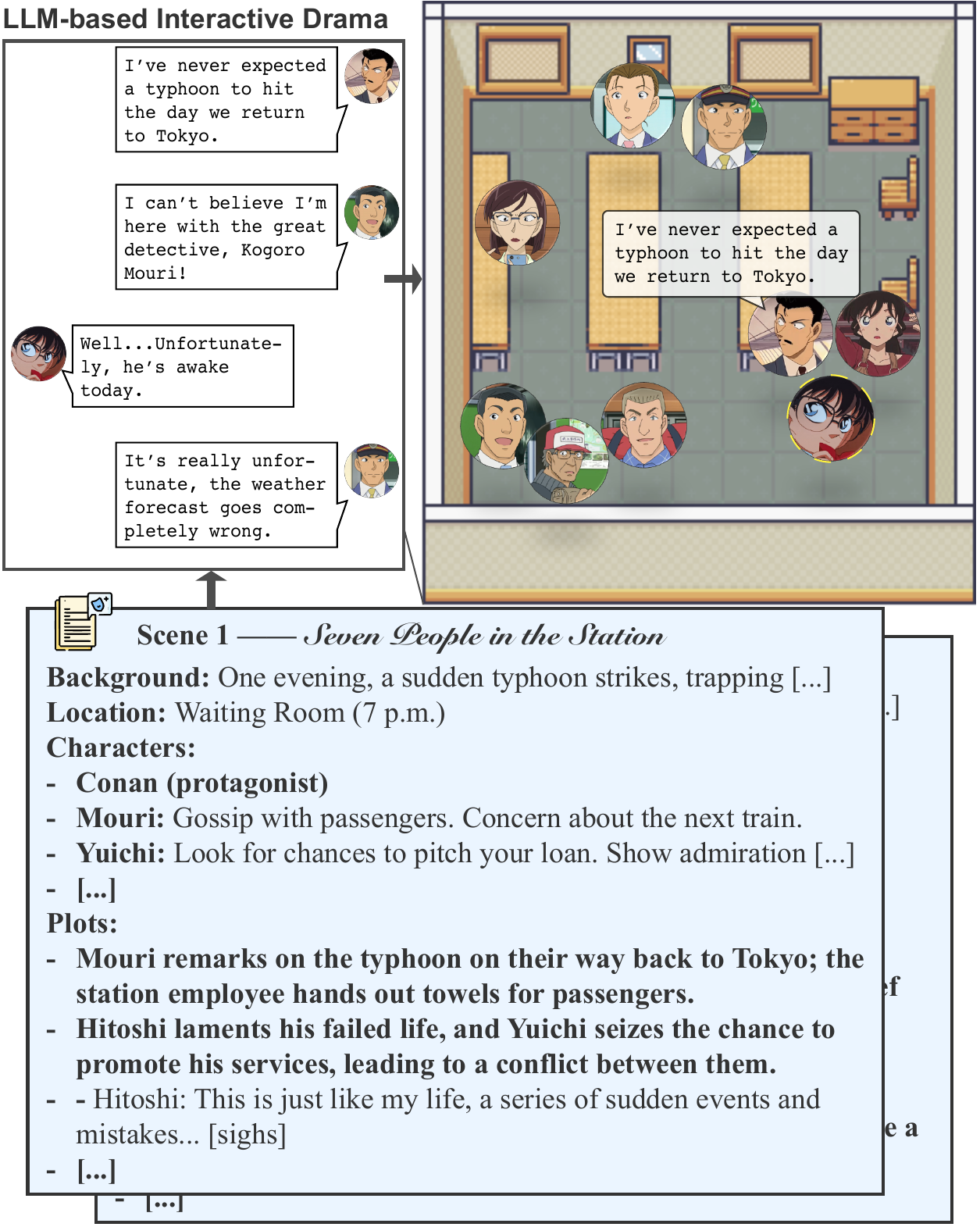}
\caption{A demonstration of LLM-based interactive drama. Rather than the imaginary scene on the left, our work studies the dialogue-form drama on the right. Below is the drama script, adapted from \textit{Detective Conan}.}
\label{fig:drama}
\end{figure}

Drama is a classical mode of storytelling and emotional expression, which utilizes dialogues to tell the story. Compared to classical drama, where the audience passively watches characters having conversations, interactive drama \citep{mateas2000neo,DBLP:conf/atal/MehtaDMM07} transfers the audience into a ``player'', who is able to interact directly with the characters and experience the story from a first-person point of view.
In LLM-based Interactive Drama, the story is performed by LLM agents \citep{DBLP:conf/nips/BrownMRSKDNSSAA20,DBLP:journals/corr/abs-2303-08774,DBLP:journals/corr/abs-2407-21783,deepseekai2025deepseekr1incentivizingreasoningcapability}.
Figure \ref{fig:drama} offers a demonstration. Guided by a drama script, the LLM agents simulate characters (e.g. \textit{Mouri}, \textit{Yuichi}) and make a conversation, while the player (playing as \textit{Conan}) interacts with them. Unlike traditional drama scripts, where every line is scripted, the drama script for LLM agents is only defined by a sequence of plots. Following the plots, they generate dialogues and instantaneous interactions with the player.
However, traditional interactive experiences are often rigid and constrained by predefined plots.
Therefore, we seek to harness LLMs' capabilities in dialogue and character simulation to break that limitation, building deeper emotional connections with the player.
In this paper, we study interactive drama in the dialogue format and do not discuss scenery generation techniques.

To understand the interactive experience enabled by LLM-based interactive drama, we introduce the two key aspects discussed in \citet{mateas2000neo}: \textbf{Immersion} and \textbf{Agency}. 
While previous work focuses on generic architectures \citep{DBLP:conf/acl/WuWJL0024,DBLP:conf/acl/HanCLXY24}, it hasn't carefully discussed these two aspects, as well as the relationship between them.

\begin{figure}[t]
\centering
\includegraphics[width=0.42\textwidth]{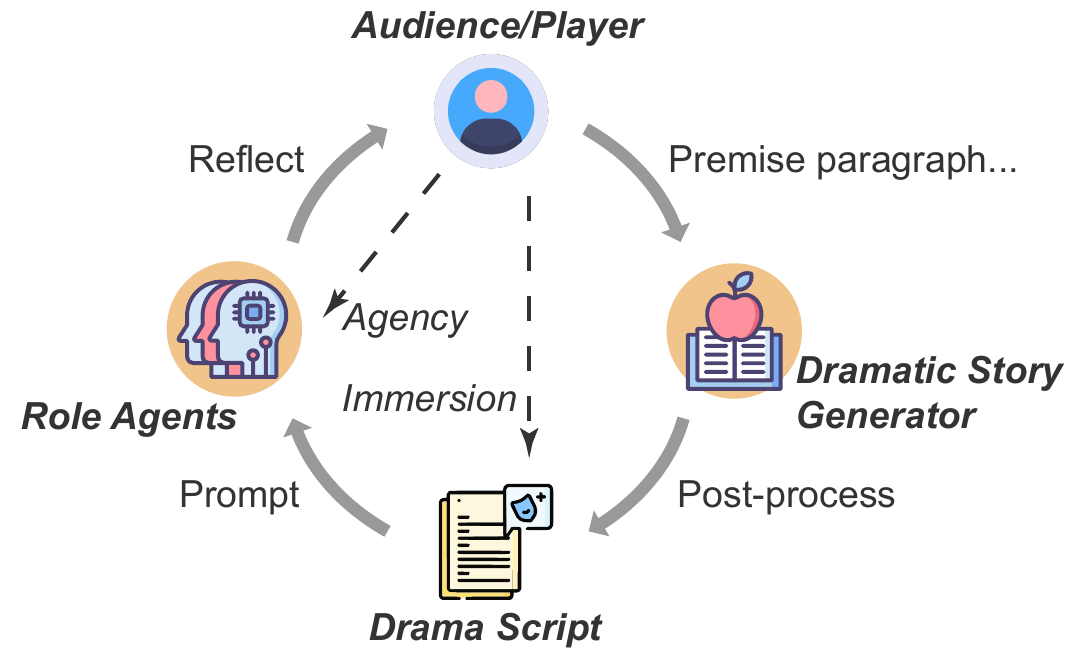}
\caption{The proposed immersion-agency paradigm for LLM-based interactive drama.}
\label{fig:loop}
\end{figure}

\textit{Immersion refers to the sensation of being present in the story.}
In addition to visual assets and music, this is fundamentally determined by the quality of the story as well as the storytelling process.
Any missing plot or interruption can break the player out of the story.
Moreover, \textit{immersion can be further enhanced when the player is able to experience the story of their own making.} This is because the self-made character may have a closer connection to the player themselves, making it easier for them to immerse in the story.

\textit{Agency refers to the ability to influence the story world.}
Agency can be straightforwardly understood as engagement. Traditionally, the player is provided with some options and specific choices can lead them to different plot branches. However, such agency has been determined by the script.
In LLM-based interactive drama, the player is endowed with more freedom by being able to have open-ended interactions with characters. We thus focus on a more general aspect of agency—the characters can exhibit meaningful transformation of reaction induced by the player's behavior. For example, a player might attempt to extract a secret from a non-player character (NPC), seek temporary companionship, or advance the plot in a specific way. The NPCs should understand the player's intentions and offer meaningful feedback.

The above theory provides a foundation for our work. We introduce the immersion-agency paradigm for LLM-based interactive drama in Figure \ref{fig:loop}.
To enhance immersion and agency based on this paradigm, we make two key contributions.

\noindent $\bullet$ The ideal vision is that players are able to create a complete drama script effortlessly and then immerse in it. This duty can be delegated to an LLM generator, which crafts the dramatic story based on a player-specified premise paragraph. However, we find the stories generated by current methods lack basic dramatic structures and compelling conflicts, leaving a significant gap to human-written ones. To mitigate this gap, we propose \textit{Playwriting-guided Generation} in Sec. \ref{sec:gen}.

\noindent $\bullet$ Previous work has largely overlooked agency in constructing LLM-based role agents.
In Sec. \ref{sec:role}, we present \textit{Plot-based Reflection}, which adapts and creates plots by analyzing player-related memories to better guide characters' behavior.

\begin{table*}[]
\centering
\small
\begin{tabularx}{0.99\textwidth}{XXXX}
\toprule
\textbf{\textit{Love}}\newline - The protagonist meets their heart's desire.\newline - A romantic relationship blooms but is soon met with obstacles.\newline - The lovers overcome the obstacles or get separated.\newline \textit{``Titanic''}
& \textbf{\textit{Phoenix}}\newline - The protagonist begins at a low point in life or society.\newline - Seize a key opportunity and face setbacks along the way.\newline - Achievement of life goals above the struggles.\newline \textit{``The Great Gatsby''}
& \textbf{\textit{Cinderella}}\newline - The protagonist is initially in a state of hardship.\newline - Receive an opportunity to escape, hindered by societal barriers like lineage.\newline - Virtue and talents are recognized towards a happy life.\newline \textit{``Jane Eyre''}
& \textbf{\textit{Love Triangle}}\newline - The protagonist is torn between two admirers.\newline - The dynamic creates competition and jealousy among the three.\newline - A choice is eventually made solving the situation.\newline \textit{``The Twilight Saga''} \\ \toprule
\textbf{\textit{Revenge}}\newline - The protagonist suffers great harm or betrayal.\newline - They devise and execute a meticulous plan.\newline - Attainment of satisfaction or a lingering sense of void.\newline \textit{``Hamlet''}
& \textbf{\textit{Family}}\newline - Complex relationships between family members.\newline - Tensions arise either from the family or external forces.\newline - Understanding and deepened affections.\newline \textit{``Little Women''}
& \textbf{\textit{Reunion}}\newline - The protagonist leaves familiar places for a reason.\newline - They grow through a series of trials and tribulations.\newline - They return home and reunite with loved ones.\newline \textit{``Odyssey''}
& \textbf{\textit{Savior}}\newline - The protagonist, faced with a responsibility or a call, decides to forge ahead.\newline - Break through difficulties and even life threats.\newline - Eventually save the people.\newline \textit{``Dune''}  \\ \bottomrule
\end{tabularx}
\caption{Eight dramatic situations. We describe them using: - Setup, - Confrontation, - Resolution.}
\label{tab:drama_sits}
\end{table*}

\section{Related Work}
Interactive drama driven by computer systems has been proposed for quite a long time \citep{DBLP:journals/sigart/BatesLR91,DBLP:books/daglib/0081917}.
More recently, \citet{DBLP:conf/acl/WuWJL0024} first discuss LLM-based interactive drama, defining its six primary elements. The director-actor architecture is first used in \citet{DBLP:conf/acl/HanCLXY24}, while \citet{DBLP:journals/corr/abs-2408-01725} model a single character with an ego and a superego to capture its development. In our work, we employ a director agent to program the plots periodically. \citet{DBLP:conf/acl/WuWJL0024} propose a curriculum learning method to fine-tune the role agents. Our work, in parallel, proposes a comprehensive framework suitable for any LLMs.

\paragraph{AI-Generated Stories}
It has been widely used in recent years to scale the training performance by using AI-generated data \citep{DBLP:conf/acl/WangKMLSKH23,DBLP:conf/emnlp/LeePSWJ23,DBLP:conf/acl/WuWJL0024}.
Our research focuses on the use of LLMs to assist humans in creating high-quality literature, such as poetry \citep{DBLP:conf/emnlp/OrmazabalAASA22,DBLP:conf/emnlp/ChakrabartyP022}, music \citep{DBLP:journals/corr/abs-2210-13944,DBLP:journals/corr/abs-2308-12982}, stories \citep{DBLP:conf/naacl/TanYAXH21,DBLP:conf/emnlp/TianHLJSCMP24,DBLP:journals/corr/Riedl16}. It is related to computational creativity \citep{DBLP:journals/ngc/Wiggins06,DBLP:journals/aim/Gervas09,DBLP:conf/ecai/ColtonW12}. Our method generates a story given a short paragraph. This is more challenging than generating the drama script given a complete story \citep{DBLP:conf/aaai/ZhaoZ0ZL0024}.

\paragraph{Simulating Dramatic Characters with LLM Agents}
Recent years emerges many companion-oriented AI applications such as \texttt{character.ai}, where LLM agents are developed to simulate various characters \citep{DBLP:journals/nature/ShanahanMR23,DBLP:conf/emnlp/ShaoLDQ23,DBLP:journals/corr/abs-2407-11484,DBLP:conf/acl/Lu0ZZ24}. A new request arises, players are not satisfied with aimless conversations, but would like to go through a story, an adventure with AI, indicating the very need for our research, which connects NLP with research in psychology \citep{DBLP:journals/corr/abs-2408-01725}, art \citep{DBLP:conf/chi/MirowskiMPE23}, and narratology \citep{todorov1969grammaire}.
Dramatic characters evolve over time, undergoing shifting inner motivations, which greatly distinguishes them from conversational agents.

\section{Playwriting-Guided Generation}
\label{sec:gen}

The dramatic story generator seeks to craft an engaging story with dramatic development based on a player-specified paragraph, which we call the premise paragraph. It includes the background, protagonist description, and beginning of the story.

However, recent studies have shown that LLMs' storytelling capabilities fall significantly short of human standards \citep{DBLP:conf/emnlp/TianHLJSCMP24,DBLP:conf/chi/ChakrabartyLAMW24}. This phenomenon is contradictory to their pre-training process, during which LLMs are exposed to a vast amount of human assets, including numerous masterpieces of literature.

Indeed, \textbf{being a brilliant playwright demands not only exposure to great works but also mastery of sophisticated playwriting techniques.} We conjecture that most LLMs' fine-tuning processes lack emphasis on the latter dimension. As a result, while LLMs accumulate a large repository of writing samples, they struggle to distill and apply the core principles from them effectively. 

In this section, we present a series of playwriting techniques and propose \textit{Playwriting-Guided Generation}, a method that integrates playwriting techniques as heuristic principles to guide the LLM generator in crafting more engaging and structured dramatic stories.
Our method turns out to significantly improve the LLMs' storytelling quality.
However, it doesn't endow LLMs with human-level power. Humans' emotions and real-life experiences enable us to create works of greater depth and resonance.

\subsection{Dramatic Situations}
\citet{polti1945thirty} identified thirty-six dramatic situations by analyzing a vast collection of literary works. Dramatic situations can be straightforwardly understood as the structure a dramatic story unfolds. More recent studies refine and condense these situations into a more streamlined set \citep{baker2022dramatic}.
These dramatic situations are more adapted to modern stories, covering vast majority of dramatic conflicts and human emotions. 
We show them in Table \ref{tab:drama_sits}. To provide a clear understanding, we illustrate each using Aristotle's three-act structure \citep{field2005screenplay}, which divides a drama into setup, confrontation, and resolution.

\subsection{Narrative Techniques}
Dramatic situations can be regarded as macro-level playwriting techniques, offering a framework for constructing a logical and efficient story structure. 
In this part, we further focus on micro-level narrative techniques. We summarize six widely-used narrative techniques, drawing from \citet{archer1913play} as well as a wide range of modern storytelling mediums (e.g. web novels, games). They are \textbf{Suspense}, \textbf{Twist}, \textbf{Non-linear Narrative}, \textbf{Multiple Narrative}, \textbf{Irony}, and \textbf{Symbolism}.
Details of each technique can be found in Appendix \ref{apx:playwriting}.

We examine the use of three representative narrative techniques in two state-of-the-art LLMs (GPT-4o \citep{DBLP:journals/corr/abs-2303-08774} and Qwen2.5-72b \citep{DBLP:journals/corr/abs-2412-15115}). Table \ref{tab:stat} compares the vanilla prompting method to our method (will be detailed below) given 50 premise paragraphs. It suggests that, without explicit prompting, both Qwen2.5 and GPT-4o are almost not able to apply any type of narrative techniques in their storytelling.

\begin{table}[]
\centering
\small
\begin{tabular}{@{}lrrr@{}}
            & \textbf{Suspense} & \textbf{Twist} & \textbf{Non-Linear} \\ \toprule
Qwen2.5-72b & 4  (8\%)          & 3 (6\%)        & 1 (2\%)             \\
GPT-4o      & 6  (12\%)         & 3 (6\%)        & 0                   \\
Our Method  & 21 (42\%)         & 37 (74\%)      & 14 (28\%)           \\
\end{tabular}
\caption{Narrative techniques used in 50 LLM-generated stories. Our method is based on GPT-4o.}
\label{tab:stat}
\end{table}

\subsection{Story Generation}
\begin{algorithm}[t]
\small
\caption{\small Playwriting-Guided Generation}
\textbf{Input:} writer LLM $\mathcal A$, critic LLM $\mathcal B$, premise paragraph $\mathbf w$, dramatic situations $\mathcal D$, narrative techniques $\mathcal N$ \\
\textbf{Output:} dramatic story $\mathcal S=\{s_k^*\}$
\begin{algorithmic}[1]
\For {$i=1$ to $3$}
    \State{Sample $1$ dramatic situation $d_i$ from $\mathcal D$}
    \State{Sample $3$ narrative techniques $n_i^{1\sim3}$ from $\mathcal N$}
    \State{Generate the story $\mathcal S_i=\mathcal A(\mathbf w,d_i,n_i^{1\sim3})$}
    \State{Comment the story $c_i=\mathcal B(\mathcal S_i,d_i,n_i^{1\sim3})$}
    \State{Revise the story by the comment $\mathcal S_i\leftarrow\mathcal A(\mathcal S_i,c_i)$}
\EndFor
\State{Vote for the best story $\mathcal S=\{s_k\}$ from $\mathcal S_{1\sim3}$}
\For {$i=1$ to $3$}
  \State{Refine the story with finer details $\{s_k^*\}\leftarrow\mathcal A(\{s_k\})$}
\EndFor
\end{algorithmic}
\label{a1}
\end{algorithm}

The entire story generation process is shown in Algorithm \ref{a1}. The generator we use is GPT-4o and we put the prompts in Appendix \ref{apx:playwriting}.
First, we initialize the sets of dramatic situations $\mathcal D$ and narrative techniques $\mathcal N$. From them, we uniformly sample one dramatic situation and three narrative techniques (line 2$\sim$3).
We represent a story $\mathcal S$ as a sequence of narrative sentences $\{s_1,s_2\cdots\}$, briefly denoted as $\{s_k\}$.
We prompt the writer LLM $\mathcal A$ to craft the story, applying the selected dramatic situation and all three narrative techniques.

\paragraph{Sampling}
The first challenge is to determine which combination of techniques best suits the given premise. We thus repeat this process multiple times (we use 3). Each iteration uses a unique dramatic situation and a unique selection of narrative techniques, resulting in three different story versions. Then, we vote for the best story by querying three additional independent LLMs (line 8).

\paragraph{Critic \& Revise}
The second challenge is that we empirically find that the writer LLM $\mathcal A$ often fails to apply all specified playwriting techniques well in its initial output story.
To address this, we introduce a critic LLM $\mathcal B$ to evaluate the output by $\mathcal A$. It is prompted to check whether $\mathcal A$ has applied the techniques properly and then offer an improvement comment.
Following the comment, $\mathcal A$ revises the story (line 5$\sim$6).
Researches suggest that external feedback helps rectify the prompt given to the LLM, thus enhancing its output \citep{DBLP:journals/corr/abs-2402-08115}.

\paragraph{Refinement}
In this paper, we focus on generating stories of about 500 words. It is a medium length compared to previous work \citep{DBLP:conf/aaai/YaoPWK0Y19,DBLP:conf/naacl/TanYAXH21,DBLP:journals/corr/abs-2201-02662,DBLP:conf/emnlp/TianHLJSCMP24}.
At this length, we observe that the generated stories are often weak in detailed presentation, e.g. plot coherence, nuanced change of character status, which is the third challenge.
To address this, we leverage the progressive generation approach \citep{DBLP:conf/naacl/TanYAXH21,DBLP:conf/nips/MadaanTGHGW0DPY23}, where the generator is prompted to refine the output text progressively from broader narrative strokes to finer details.
Specifically, we continually prompt the writer LLM to add details to existing narrative sentences or insert new sentences between them (line 10), until we achieve the ultimate story $\{s_k^*\}$.

\subsection{Transfer Story To Drama Script}
The last step is to post-process the story $\mathcal S$ into a standard drama script, using GPT-4o.
As shown in Figure \ref{fig:drama}, a drama script follows an episodic structure.
Each scene has an independent character setup as characters' thoughts can shift throughout the story.
In addition, each scene contains a sequence of plots, detailing the story's progression.
The post-processing first segments the story into 3$\sim$5 scenes. Specifically, for each scene, it extracts and adjusts the narrative sentences from $\mathcal S$ into a sequence of plots. Here, any flashback and flashforward should be treated as an independent scene. Then, it crafts the background, location, and character setup. Note that this process introduces no creativity job; rather, it processes information directly from the generated story.

\section{Role Agents}
\label{sec:role}

This section explores the construction of LLM-based role agents. Their overall duty is to simulate the dramatic characters and progress the storytelling, following the drama script.

\subsection{Preliminary}
Let us regard the entirety of role agents as a black-box decision function ${\rm Decision}$. At each moment $t$, it processes some text inputs and makes a decision (e.g. speak something to someone) $y_t={\rm Decision}(x_t,m_t,o_t,\sigma)$, where $x_t$, $m_t$, $o_t$, and $\sigma$ denote the player input, memories, observations, and drama script, respectively.

To ensure the story develops as defined, previous work introduces the \textbf{plot chain} \citep{DBLP:conf/acl/WuWJL0024}, a sequence of plot objectives that the player and characters should progress through. The role agents will preserve this plot chain throughout the story in addition to making the decision.
Hence, the decision step can be formulated as:
\begin{equation}
    p_{t+1},y_t={\rm Decision}(x_t,p_t,m_t,o_t,\sigma)
    \label{eq:decision}
\end{equation}
where $p_t$ and $p_{t+1}$ denote the current and next state of the plot chain, In a plot chain, complete plots will be tagged as ``true'' and incomplete plots will be tagged as ``false''.
Note that the process in Eq. \ref{eq:decision} is done within one inference. We first update the state of the plot chain and then make the decision.

In addition, to handle the player's provocative and off-track inputs, \citet{DBLP:conf/acl/WuWJL0024} leverage some replying strategies to guide the player back to the plot. We further diversify the strategies to ensure a more attractive way of guidance (in Appendix \ref{apx:replying}).

\subsection{Plot-based Reflection}
Dramatic characters are shaped not only by their profiles but more importantly by their thoughts (i.e. inner motivations), which drive their dramatic development across different scenes. Our vision is that \textbf{player agency can be greatly expressed by the adaptive dynamics of character's inner motivations}. To do this, we propose \textit{Plot-based Reflection}. This mechanism enables the role agents to adapt the plots by analyzing player's behavior (e.g. emotion, intention), to better motivate the characters' reactions. The reflection step is formulated as:
\begin{equation}
    p_t^*={\rm Reflection}(x_t,p_t,m_t,o_t,\sigma)
    \label{eq:reflection}
\end{equation}
where $p_t^*$ denotes the reflected version of plot chain. After the reflection step, the decision step is followed as in Eq. \ref{eq:decision}, ensuring that the story progresses following $p_t^*$.
Plot-based reflection will be made periodically every $k$ moment. We set $k=5$. If $k$ is too small, there won't be enough contextual information to adapt plots. If $k$ is too large, there will be no agency.

Unlike memory-based reflection \citep{DBLP:conf/uist/ParkOCMLB23}, which synthesizes new memories based on past memories, plot-based reflection adapts plots based on player's behavior in memories. Two are parallel techniques during the agents' decision-making process. Concrete examples of plot-based reflection can be found in Table \ref{tab:case}.

Moreover, plot-based reflection should be bounded.
Our approach is explicitly restricting one reflection step to adjust no more than one incomplete plot or insert no more than one new plot.
We find that the LLM is inclined to adjust the current plots to a greater extent, thus introducing plots that are incoherent or unrelated to the original plots.
We will further discuss this issue in the following qualitative analysis.

\subsection{Architecture}
\begin{figure}[t]
\centering
\includegraphics[width=0.42\textwidth]{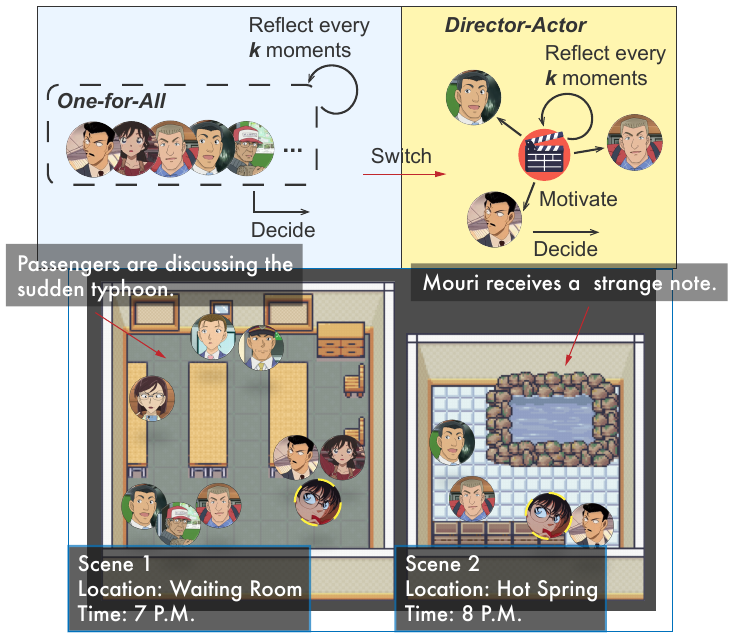}
\caption{ and plot-based reflection. The first scene emphasizes background telling. We apply the one-for-all architecture. The following scenes involve mutual suspicion and personal investigation. We apply the director-actor architecture.}
\label{fig:hybrid}
\end{figure}

We consider plot-based reflection into two primary role agent architectures, shown in Figure \ref{fig:hybrid}.

\paragraph{Director-Actor}
This is a multi-agent architecture, where each character is modeled by an independent actor agent and there is a higher-level agent (i.e. the director), which oversees and coordinates all actor agents from a global perspective. Specifically, at each moment, the director agent instructs a specific actor agent by generating a motivation $z_t$ for it, denoted as the motivation step: $p_t,z_t={\rm Motivation}(x_t,p_t,m_t,o_t,\sigma)$.
On the other hand, the actor agent won't be exposed to the script; rather, its decision is driven by its character profile $\rho$ and motivation $z_t$ from the director, $y_t={\rm Decision}(x_t,z_t,m_t,o_t,\rho)$.

We let the director agent to reflect. This is because it can see broader and farther compared to actor agents, allowing it to make the optimal solution. Its reflection step is the same as Eq. \ref{eq:reflection}.

The director-actor architecture allows each character to process its own memory and think independently, providing advantages for agency yet a greater inference complexity.

\paragraph{One-for-All}
Compared to the director-actor architecture, a simplified and lightweight alternative is to use a single global agent to role-play all characters as well as their interplay. We denote this architecture as ``One-for-All''. At each moment, the global agent chooses one character in the scene and directly makes the decision for it, bypassing the motivation step.
Based on this architecture, the reflection step and the decision step are exactly the same as Eq. \ref{eq:reflection} and Eq. \ref{eq:decision}.

\paragraph{Hybrid Architecture}
A normal decision-making without plot-based reflection in the director-actor architecture necessitates twice of inference, while the one-for-all necessitates only one.
While it greatly boosts computational efficiency, it is generally limited in the depth of characters' responses, leading to a lower sense of agency. However, it could be an economical alternative for scenes that emphasize narrative.
Not every scene in a drama encourages heavy player agency. For instance, some expository scenes that establish the story background naturally offer limited opportunities for interaction, where immersion is more critical than agency.

To balance performance and efficiency, we apply a , which dynamically switches between two architectures based on the characteristic of the scene, as shown in Figure \ref{fig:hybrid}. For scenes that emphasize interaction (e.g. exploring clues), the director-actor architecture remains the optimal choice. For scenes that emphasize narrative, we simplify it to one-for-all, which means the director agent reflects and makes the decision directly.

\subsection{Other Implementation Details}
There are other important details to implement the role agents.
$\bullet$ Model: As in previous work \citep{DBLP:conf/uist/ParkOCMLB23,DBLP:conf/aaai/ZhaoZ0ZL0024}, we prompt the state-of-the-art LLM to construct agents. In this work, the director and actor agents are all based on GPT-4o. It is a general LLM instead of reasoning LLM like OpenAI-o1 and DeepSeek-R1 \citep{deepseekai2025deepseekr1incentivizingreasoningcapability}.
$\bullet$ Memory: We retain and flatten all memories in the prompt.
$\bullet$ Transition: A scene will transit to the next when all plots in the plot chain have been completed.

\begin{table*}[]
\centering
\small
\setlength\tabcolsep{8.5pt}
\begin{tabular}{lccccccccc}
                                     & \multicolumn{2}{c}{\textsc{Conflict}} & \multicolumn{2}{c}{\textsc{Suspense}} & \multicolumn{2}{c}{\textsc{Emotion}} & \multicolumn{2}{c}{\textsc{Character}}  & \textsc{Follow} \\  \cmidrule(lr){2-3} \cmidrule(lr){4-5} \cmidrule(l){6-7} \cmidrule(l){8-9}
\textsc{Method}                      & B$\uparrow$ & W$\downarrow$ & B$\uparrow$ & W$\downarrow$ & B$\uparrow$ & W$\downarrow$ & B$\uparrow$ & W$\downarrow$ \\ \toprule
\textit{Outline-First}               & \underline{18\%} & \underline{34\%} & \underline{10\%} & 28\%             & \underline{10\%} & \underline{50\%} & 18\%             & \underline{36\%} & /    \\
\textbf{\textit{Playwriting-Guided}} & \textbf{32\%}    & 24\%             & \textbf{32\%}    & \textbf{22\%}    & \textbf{48\%}    & 16\%             & 34\%             & 20\%             & \textbf{92\%} \\
\quad\textit{w/o. critic \& revise}  & 24\%             & 24\%             & 26\%             & \underline{34\%} & 18\%             & 28\%             & \underline{12\%} & 32\%             & 66\% \\
\quad\textit{w/o. refinement}        & 26\%             & \textbf{18\%}    & 32\%             & 26\%             & 24\%             & \textbf{6\%}     & \textbf{36\%}    & \textbf{12\%}    & /    \\
\end{tabular}
\caption{Win-rates of different dramatic story generation methods over 50 topics. We use the win-rate ``B'' to represent the ratio that a method performs the best and use the win-rate ``W'' to represent the probability that a method performs the worst. We highlight the best and worst score using \textbf{bold} and \underline{underline} for each dimension.}
\label{tab:story}
\end{table*}

\section{Experiment}
Our experiments are evaluated by human annotators. 
LLMs have been shown to be able to take the duty on some assessment work for humans \citep{DBLP:journals/corr/abs-2407-11484}. Despite the high efficiency, studies show that LLM evaluators lag significantly behind human evaluators in terms of accuracy and empathy \citep{DBLP:journals/corr/abs-2303-04048}, which are the key requirements for evaluating literary works.

We employ six annotators. All of them are native speakers with a background in humanities and are exposed to dramatic concepts and theories. Four of them are undergraduates and two of them are graduate students.

\subsection{Evaluation on Dramatic Story Generation}

\begin{figure}[t]
\centering
\includegraphics[width=0.49\textwidth]{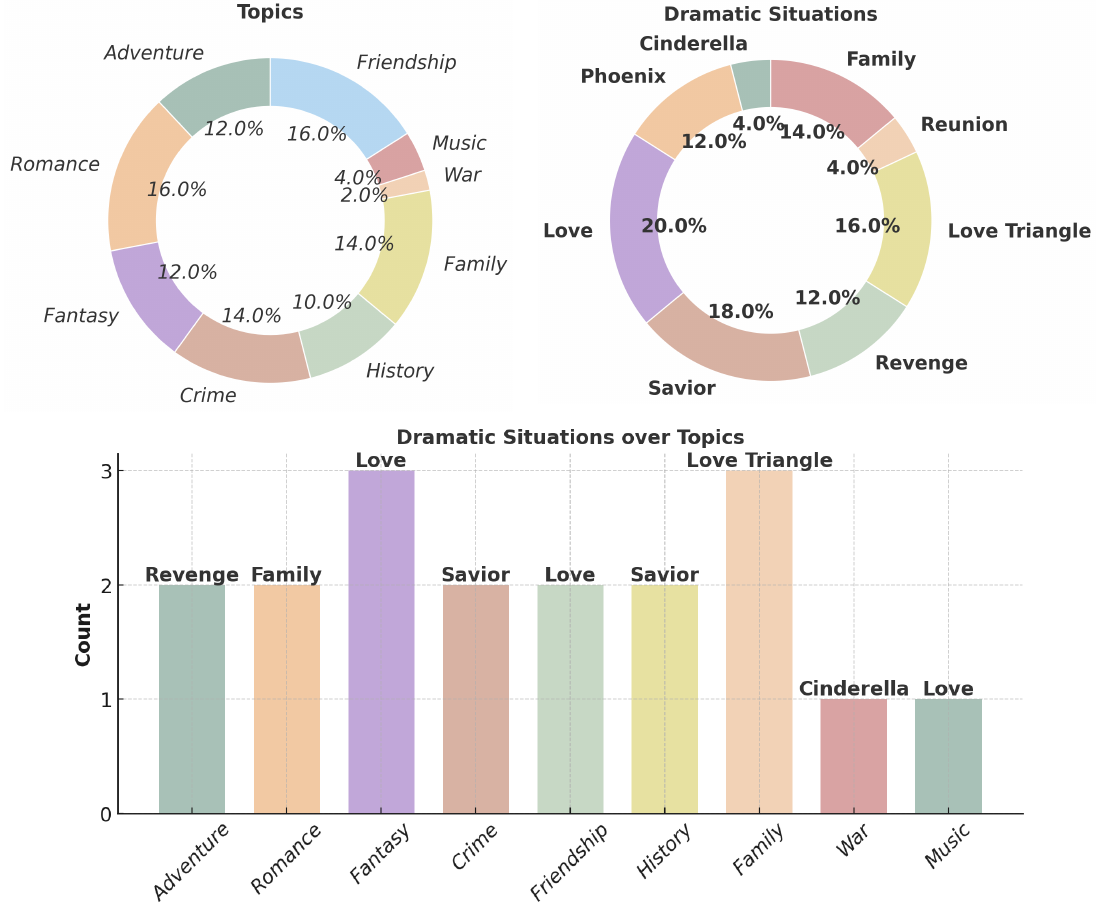}
\caption{Distribution of topics and dramatic situations generated by the algorithm.}
\label{fig:story_analysis}
\end{figure}

\paragraph{Data Preparation}
We prepare 50 carefully crafted premise paragraphs for this part of the experiment.
Our annotators, all with backgrounds in drama, manually craft the paragraphs from a diverse range of topics. To mitigate the risk of LLMs' memorization, we ask the annotators to avoid using names or symbols that are popular in modern literature. The distribution of the topics are shown in Figure \ref{fig:story_analysis}. Each paragraph contains 50$\sim$100 words.

\paragraph{Main Results}
We compare with outline-first prompting \citep{DBLP:conf/emnlp/TianHLJSCMP24}, a stronger baseline over vanilla prompting, where the LLM first writes an outline before crafting the complete story. Note that in our method, we also prompt the LLM to write an outline first.
Table \ref{tab:story} reports the win-rates from four dimensions, \textbf{conflict}, \textbf{suspense}, \textbf{emotional tension}, and \textbf{character arc}.
All dimensions gain significantly from ``playwriting-guided'', particularly pronounced in suspense (10\%$\rightarrow$32\%) and emotional tension (10\%$\rightarrow$48\%) over ``outline-first''. Additionally, we find that refinement contributes greatly to emotional tension (24\%$\rightarrow$48\%). This is because emotional tension often emerges from nuanced details within the text, which can be better developed through a progressive process, while the conflict and suspense have been largely determined by the structure.
We also calculate the percentage of stories that follow the selected techniques, suggesting the effectiveness of introducing a critic LLM.
Playwriting-guided generation trades efficiency for quality harnessing multiple agents, taking 10$\sim$12 times longer than vanilla prompting.

\paragraph{Sampling Effectiveness}
The effectiveness of selecting suitable combinations of playwriting techniques is a critical factor for performance. Figure \ref{fig:story_analysis} visualizes the selected dramatic situations across different topics. We find that \textit{Romance}, \textit{Family}, and \textit{Crime} are the three most frequent topics in the premise paragraphs. Notably, all of them are highly matched with \textit{Love}, \textit{Savior}, and \textit{Family}, the three most frequent dramatic situations in generated stories.
Furthermore, at the bottom of Figure \ref{fig:story_analysis}, we list a detailed count of the most frequent dramatic situation selected for each topic. All results suggest that our algorithm samples appropriate playwriting techniques given premise paragraphs effectively.

\subsection{Evaluation on Role Agents}

\begin{table*}[]
\centering
\small
\begin{tabular}{lcccccccc}
                                      & \multicolumn{4}{c}{\textsc{Immersion}} & \multicolumn{2}{c}{\textsc{Agency}} \\ \cmidrule(lr){2-5} \cmidrule(l){6-7}
\textsc{Architecture}                 & Consistency  & Attract      & Complete     & Progress     & Influence    & Intention    & Speedup \\ \toprule
Director-Actor                        & 3.9          & \textbf{4.2} & 3.8          & 3.6          & \textbf{4.2} & 3.9          & x1.00   \\
\textbf{}          & \textbf{4.1} & 3.9          & \textbf{4.3} & \textbf{4.3} & 4.0          & \textbf{4.0} & x1.49   \\
\quad w/o. \textbf{Plot-based Reflection} & 4.0          & 3.5          & 4.2          & 3.9          & 3.5          & 3.3      & x1.90   \\
\end{tabular}
\caption{Multi-dimension evaluation of different role agent architectures. We ask the annotators to rate each dimension using an integer score from 1 to 5 where 5 represents the best and 1 represents the worst.}
\label{tab:arch}
\end{table*}

We evaluate role agents based on a high-quality manual script \textit{Seven People In the Station}, adapted from \textit{Detective Conan}.
This interactive drama has three scenes. The player plays as \textit{Conan} and there are other eight NPCs. Further details of this script are in Appendix \ref{apx:demo}.

\paragraph{Human/LLM as Player}
We invite 10 volunteers to experience our interactive drama. Before starting, we ask them to create a player personality for themselves, which could be either the same as their own or any other one they prefer. After a negotiation, each volunteer is given a unique personality, which they would follow throughout the playing. Our goal is to capture a broad range of player characteristics.
On the other hand, we consider utilizing LLM agents as players to further broaden the player base.
Particularly, we focus on those aggressive personalities, which may pose greater challenges to the system robustness.
We construct 10 agents based on GPT-4o, e.g. a grumpy guy (actions show impatience), a fan girl (always raising irrelevant topics), detailed in Appendix \ref{player_agents}.

\paragraph{Metrics}
We quantify immersion using four dimensions.
$\bullet$ \textbf{Character Consistency}: This dimension assesses how well the characters' responses align with their setup.
$\bullet$ \textbf{Character Attractiveness}: It is an advanced and comprehensive dimension, encompassing aspects like humor and empathy displayed during the interaction with the player.
$\bullet$ \textbf{Narrative Completeness}: It is critical that the player is able to experience the complete story.
$\bullet$ \textbf{Narrative Progression}: It assesses the pacing of the plot development.

For agency, we use two dimensions.
$\bullet$ \textbf{Player Influence}: It assesses how player's behavior influences the course of the story.
$\bullet$ \textbf{Intention Following}: Beyond being influenced, it assesses how characters' reactions fulfill player's intentions.

\paragraph{Main Results}
Table \ref{tab:arch} compares the performance of different architectures. For each architecture, we let each human/agent experience the interactive drama once. This results in 60 playing histories totally. We report the average score of each dimension. The  refers to applying one-for-all for the first scene and director-actor for the second and third scene.
We first focus on the effect of plot-based reflection. It significantly enhances agency from both dimensions (3.5$\rightarrow$4.0 and 3.3$\rightarrow$4.0). We also notice that plot-based reflection is useful in enhancing character attractiveness (3.5$\rightarrow$3.9) and narrative progression (3.9$\rightarrow$4.3). We conjecture that the reflection encourages characters to display a stronger sense of empathy, which thus contributes to higher attractiveness and decent progression.

The last column of Table \ref{tab:arch} suggests the efficiency boost achieved by the . We find that the director-character architecture is advantageous in building character attractiveness because the independent thinking process is more likely to give impressive responses. However, it incurs a lower quality of narrative, especially in progression (4.3$\rightarrow$3.6). This is due to the fact that the communication between director and actor agents inevitably results in information loss.

\paragraph{Human Player vs. Aggressive Agent Player}
Table \ref{fig:player_analysis} further indicates that the  remains decent in the face of two types of players. Interestingly, it even higher scores in character attractiveness and player influence against aggressive agents compared to human players. We conjecture that this is because agents are more eager to interact and the high-quality responses in turn impress the annotators. However, the provocative queries might lead to missing of plots, as shown by the drop in narrative progression.
\begin{table}[]
\centering
\small
\begin{tabular}{@{}lccc@{}}
                          & Attractive & Progression & Influence \\ \toprule
\textit{Human}            & 3.8        & 4.5         & 3.9       \\
\textit{Aggressive Agent} & 4.0        & 4.1         & 4.1       \\
\end{tabular}
\caption{Comparing humans and aggressive agents.}
\label{fig:player_analysis}
\end{table}

\begin{table}[]
\centering
\tiny
\begin{tabular}{@{}ll@{}}
\toprule
\textbf{Key Memories}                      & \textbf{Reflected Plot Chain} \\ \midrule
\begin{tabular}[c]{@{}l@{}}\textbf{Conan:} Uncle, what\\ happened to your\\ life?\end{tabular} & \begin{tabular}[c]{@{}l@{}}- Hitoshi laments his failed life, and Yuichi seizes the chance\\ to promote his business, leading to a conflict between them.\\ \hlblue{- Conan grows curious about Hitoshi, prompting Hitoshi to}\\ \hlblue{reluctantly give him a brief account of his factory's closure.}\\ - Yuichi's suitcase falls open [...]\end{tabular} \\ \midrule

\begin{tabular}[c]{@{}l@{}}\textbf{Conan:} Is there\\ someone in this\\ room that makes\\ you scared?\end{tabular} & \begin{tabular}[c]{@{}l@{}}- Apologize to Noriko and promise to recreate the poster\\ tomorrow.\\ \hlblue{- Cryptically convey the message of the note to Conan.}\\ - Provide towels for passengers.\end{tabular} \\ \midrule

\begin{tabular}[c]{@{}l@{}}\textbf{Conan:} Miss\\ Masako, you don't\\ look well.\\ I'd love to help you.\end{tabular} & \begin{tabular}[c]{@{}l@{}}- The stationmaster lets everyone free to use the station's\\ facilities; Morris suggests going to the hot spring together.\\ \hlpurple{- Masako secretly hands Conan a note: ``There is danger here.''}\\ \textbf{\#\# Script leakage}\end{tabular} \\ \midrule

\begin{tabular}[c]{@{}l@{}}\textbf{Conan:} Does\\ anyone here like to\\ play tennis?\end{tabular} & \begin{tabular}[c]{@{}l@{}}- Mouri remarks on the typhoon on their way back to Tokyo [...]\\ \hlpurple{- The stationmaster gives a brief introduction to the station's}\\ \hlpurple{activity areas, mentioning a hot spring and a small tennis court.}\\ - Hitoshi laments his failed life, and Yuichi seizes the [...]\\ \textbf{\#\# A scene that doesn't exist}\end{tabular} \\ \bottomrule

\end{tabular}
\caption{Qualitative analysis of plot-based reflection. We highlight the parts that are adapted by the reflection.}
\label{tab:case}
\end{table}

\paragraph{Qualitative Analysis}
Table \ref{tab:case} showcases the qualitative analysis probing plot-based reflection. In the second case, Masako gradually opens up after the player's skillful probing, agreeing to convey the secret to him. In addition to good cases (the first and second), we particularly focus on two bad cases that highlight the direction for improvement. In the third case, the reflection correctly responds to player's interest in Masako. However, it brings up a key plot from the next scene. This is because the global agent responsible for reflection has access to the entire script, leading to unintended information leakage. In the fourth case, the new plot introduces a nonexistent scene, the tennis court, in response to player's input. Consequently, it can be problematic if the player attempts to explore it.

These issues can be resolved by alignment of the model (SFT, RLHF) within the close-domain story world. We will leave this for our future work.

\section{Conclusion}
This paper enhances immersion and agency for LLM-based interactive drama by optimizing the story generation and role agent architecture. Our evaluation relies on human judgment instead of LLM judgment. We hope our work contributes to advancements in novel dialogue and game systems.

\section*{Limitations}
Our paper adopts a limited scope of playwriting techniques. However, in modern drama, they should not be predetermined; they should evolve and be adapted based on various domains.
There is a potential gap between two parts of evaluation. We evaluate dramatic story generation based on 50 manually-crafted premises, while role agents are evaluated based on a high-quality script.
We have not found an efficient method to quantify the player's complete experience across both the creation and engagement of the story. For example, inviting the volunteers to craft the story, play the interactive drama, and eventually make comments can be inefficient and unstable.
For our demonstration, we develop visual scenery using the Phaser3 game engine. However, the drama we study in this paper focuses on text-based interactive experiences.
Our method is based on prompting GPT-4o, one of the most powerful LLMs, but it still exhibits shortcomings in plot-based reflection and making ideal character decisions. Our future work will focus on developing effective data-driven methods to enhance the specific capabilities.
The memory system we use in constructing the LLM agents is out-of-date. We will further improve this aspect.



\bibliography{custom}

\appendix
\newpage

\section{Playwriting-Guided Generation}
\label{apx:playwriting}

We detail the narrative techniques below.

\noindent$\bullet$ \textbf{Suspense}:
It usually occurs at the beginning of a story, sparking the audience's curiosity by deliberately creating an information mismatch, e.g. hiding some key details.

\noindent$\bullet$ \textbf{Twist}:
It is a widely-used trick to set up a twist outside the audience's expectations during the plot development to create powerful dramatic impact. Some works include multiple twists to further intensify the impact.

\noindent$\bullet$ \textbf{Non-Linear Narrative}:
Break free from chronological storytelling by rearranging the timeline, e.g. flashbacks (reveal past events), flashforwards (offer glimpses into the future).

\noindent$\bullet$ \textbf{Multiple Narrative}:
The audience plays the role of multiple characters and explores the story through distinct lenses, offering a more comprehensive view of the story.

\noindent$\bullet$ \textbf{Irony}:
Intentionally contrast a character's words or actions with the actual situation to create humor and sharp criticism.

\noindent$\bullet$ \textbf{Symbolism}:
Employ symbolic objects or imagery to convey deeper or abstract messages. For instance, doves symbolize peace.

Table \ref{tab:playwriting_prompts} demonstrates the prompts for generating, commenting, revising, and refining the story.

\section{Replying Strategies}
\label{apx:replying}

To guide the player back to the plot, each character performed by an LLM agent is prompted by specific replying strategies in response to a wide variety of player inputs.
We notice that some player inputs contribute meaningfully to the storytelling, making it more engaging and enhancing immersion. However, some disruptive or provocative behavior can destroy the storytelling. Therefore, we first classify the player input into three categories before replying: $\bullet$ \textbf{In-plot}: the input aligns with the ongoing plot; $\bullet$ \textbf{Daily}: the input is outside the plot, but relevant to the story and makes sense within the context; $\bullet$ \textbf{Breaking}: the input will break the storytelling, either because it is irrelevant to the plot, nonsensical, provocative, or even offensive, etc.

Based on the judgement, in the face of two out-of-plot inputs (daily and breaking), the characters should guide the player back to the plot in a logical and decent way, neither spoiling the performance nor frustrating the player.
To do this, we present three strategies.

\noindent$\bullet$ \textbf{Avoid}: Simply avoid the irrelevance of what the player says and then redirect the conversation back on track. For instance:

\noindent\textit{Conan}: \textit{Uncle Mouri, I hear you're a good code writer.}

\noindent\textit{Kogoro}: \textit{You little brat, what nonsense do you talk all day long, don't come to disturb the adults working on the case, or I'll whack you.}

\noindent$\bullet$ \textbf{Ignore-Question}: The most straightforward strategy is to pretend not hearing what the player says and talk about something else. However, a outright ignoring can be somewhat rude and a loss of fun. A takeaway is the character turns the table and initiates a question to the player. This is a useful technique to bring the player and the character personae closer together, enhancing immersion. For instance:

\noindent\textit{Conan}: \textit{Uncle Mouri, I hear you're a good code writer.}

\noindent\textit{Kogoro}: \textit{Kid, do you think the criminal is among the three of them?}

\noindent$\bullet$ \textbf{Associate}: A blanket avoidance can come across as very mechanical on the part of the character. A golden strategy would be to associate some extraneous entity or imagery in the player input to one that is relevant to the plot. Such responses would create smooth turns in dialogue and make the character more humorous and attractive.
For instance:

\noindent\textit{Conan}: \textit{Uncle Mouri, I hear you're a good code writer.}

\noindent\textit{Kogoro}: \textit{You little brat... Couldn't be saying that the victim left some message at the scene.}

\noindent Compared to the first two, this strategy is the most attractive, which heavily relies on the LLM's understanding of the character and world knowledge.

\onecolumn
\small
\begin{longtable}{p{\linewidth}}
\toprule
\textbf{Story Generation} \\
\midrule

\textbf{\#\# Task}\\
Create a dramatic story from the protagonist's perspective based on the premise paragraph. The premise paragraph may include the background of the story, the protagonist, the beginning, and the ending. \\Interactive drama differs from traditional drama in that the audience also plays a character in the story, namely the protagonist, and can interact with the characters in the drama to experience the story firsthand. \\\\
\textbf{\#\# Premise Paragraph}: \\\{topic\} \\\\
\textbf{\#\# Dramatic Situation}: \\\{situation\} \\\\
\textbf{\#\# Narrative Techniques}: \\\{techniques\} \\\\
\textbf{\#\# Workflow}: \\
1. The creation must follow the specified dramatic mode, and all the techniques listed in dramatic techniques must be used. \\
2. First, provide introductions to the main characters in the drama other than the protagonist, with each character description being around 100 words. Also, give a brief introduction to the protagonist. \\
- Characters must be specific individuals; vague roles such as "classmates around" or "audience" are not allowed. \\
3. You should firstly create a plot outline, around 100 words. \\
4. Expand the plot outline into a complete story, around 500 words. \\
- Note: If multiple narratives are used, the protagonist will temporarily switch to another character. \\\\
\textbf{\#\# Output Format}: \\
\#\#\# Plot Outline \\
\#\#\# Complete Story \\
\#\#\# Technique Explanation (briefly explain how the techniques are reflected in the story) \\

\hdashline

\textbf{Dramatic Situation}:\\
\textbf{Love}:\\ - The protagonist meets their heart's desire.\\ - A romantic relationship blooms but is soon met with obstacles.\\ - The lovers overcome the obstacles or get separated.\\
\textbf{Phoenix}:\\ - The protagonist begins at a low point in life or society.\\ - Seize a key opportunity and face setbacks along the way.\\ - Achievement of life goals above the struggles.\\
\textbf{Cinderella}:\\ - The protagonist is initially in a state of hardship.\\ - Receive an opportunity to escape, hindered by societal barriers like lineage.\\ - Virtue and talents are recognized towards a happy life.\\
\textbf{Love Triangle}:\\ The protagonist is torn between two admirers.\\ - The dynamic creates competition and jealousy among the three.\\ - A choice is eventually made solving the situation.\\
\textbf{Revenge}:\\ - The protagonist suffers great harm or betrayal.\\ - They devise and execute a meticulous plan.\\ - Attainment of satisfaction or a lingering sense of void.\\
\textbf{Family}:\\ - Complex relationships between family members.\\ - Tensions arise either from the family or external forces.\\ - Understanding and deepened affections.\\
\textbf{Reunion}:\\ - The protagonist leaves familiar places for a reason.\\ - They grow through a series of trials and tribulations.\\ - They return home and reunite with loved ones.\\
\textbf{Savior}:\\ - The protagonist, faced with a responsibility or a call, decides to forge ahead.\\ - Break through difficulties and even life threats.\\ - Eventually save the people.\\

\hdashline

\textbf{Narrative Techniques}:\\
\textbf{Suspense}: Often appears at the beginning of a story, deliberately creating an information imbalance. For example, by withholding key information, it piques the audience's curiosity and encourages them to continue seeking answers.\\
\textbf{Twist}: Introduces unexpected turns in the plot or character development, breaking the audience's preconceptions and creating a sense of shock. For example, the protagonist discovers they have been betrayed, or a character who was thought to be an ally is revealed to be the antagonist.\\
\textbf{Non-linear Narrative}: Disrupts the chronological order of the story through techniques such as flashbacks and flash-forwards. Flashbacks reveal events from the past, while flash-forwards show future events, adding complexity and depth to the narrative.\\
\textbf{Multiple Narrative}: The protagonist takes on different roles in various settings, and the story is told through multiple perspectives. This technique provides a more comprehensive and intricate narrative by showing different facets of the story.\\
\textbf{Irony}: Intentionally presents characters or statements in a way that contrasts with the actual situation, creating humor, satire, or a deeper critique. For example, a character might say something that is the opposite of what they truly mean or what is happening.\\
\textbf{Symbolism}: Uses symbolic objects or characters to represent deeper concepts or themes. For example, a dove often symbolizes peace, while a raven can symbolize bad luck or, in some contexts, power.\\

\toprule
\textbf{Story Critique} \\
\midrule

\textbf{\#\# Task}: \\Evaluate the given dramatic story from the perspective of dramatic techniques and provide comment for improvement. \\\\
\textbf{\#\# Narrative Techniques}: \\\{techniques\} \\\\
\textbf{\#\# Story}: \\\{story\} \\\\
\textbf{\#\# Requirements}: \\Understand the given dramatic techniques, identify which techniques are used in the story, up to three. Evaluate whether these techniques are applied effectively, for example, whether the twist has an impact, or if the non-linear narrative has separate scenes. Provide comment for improvement. \\\\
\textbf{\#\# Output Format}: \\
Dramatic techniques used... \\
Effectiveness of the techniques... \\
Comment... \\

\toprule
\textbf{Story Revision} \\
\midrule

\textbf{\#\# Task}: \\Revise the story based on the comments provided. \\\\
\textbf{\#\# Requirements}: \\Carefully analyze the comments provided. You can revise the scenes or the characters. \\\\
\textbf{\#\# Comment}: \\\{comment\} \\\\
\textbf{\#\# Output Format}: \\
\#\#\# New Story \\
\#\#\# Explanation (briefly explain the improvements made to the new story) \\

\toprule
\textbf{Story Refinement} \\
\midrule

\textbf{\#\# Task}: \\Refine the plot of the story. \\\\
\textbf{\#\# Requirements}: \\
Refine the plot from multiple dimensions.\\
- Coherence: Analyze the logical relationships between narratives, and modify or add new narrative sentences to enhance coherence.\\
- Detail: Find out what is not specific enough in each plot and refine these details, no limit on the number of words; if there is suspense in the narrative process, portray the suspense finely.\\\\

\textbf{\#\# Output Format}: \\
\#\#\# Analysis (analyze multiple dimensions of the current story) \\
\#\#\# Refined Story \\

\bottomrule
\caption{Prompts for playwriting-guided generation.} 
\label{tab:playwriting_prompts}
\end{longtable}

\section{Demonstration}
\label{apx:demo}

\begin{longtable}{p{\linewidth}}
\toprule
\multicolumn{1}{c}{\textbf{Characters}} \\
\midrule
\textbf{Conan} \\
A first-grade elementary student living with Kogoro Mouri and Ran Mouri. Maintains a childlike appearance while hiding exceptional deductive abilities. \\

\textbf{Kogoro Mouri} \\
A renowned detective with comedic flaws - appears lecherous, sloppy, and money-oriented on the surface. Possesses strong sense of justice and occasionally displays brilliant deduction when loved ones are involved. Father of Ran Mouri. \\

\textbf{Ran Mouri} \\
A high school student with gentle demeanor. Demonstrates remarkable courage and martial arts skills in crises. Daughter of Kogoro Mouri. Secretly misses her childhood friend Shinichi Kudo. \\

\textbf{Yuichi Hokari} \\
A fraudulent loan salesman posing as friendly, talkative and proactive consultant, showing admiration for Kogoro Mouri. He often offers help or suggestions to others voluntarily, but his ultimate goal is to find opportunities to sell loans. His true identity is that of a fraudster. He carries numerous brochures from different companies in his briefcase and does not want anyone to have a chance to look inside it.

\textbf{Carl Morris} \\
An American backpacker with ulterior motives. Skilled pickpocket using friendly persona to gain trust. Appearing to be enthusiastic, cheerful, and helpful, loving traveling and learning languages. Attempts to steal valuables during the lockdown. \\

\textbf{Hitoshi Takegami} \\
A bankrupt factory owner driven to despair. His business failure has led him to be pessimistic and angry, and he harbors a strong dislike for salespeople. Was once swindled by a loan salesman and forced to close his factory due to usurious loans from the bank. Carries a modified pistol with thoughts of suicide but hesitating to go through with it. \\

\textbf{Noriko Kurusu} \\
Secretary to corrupt councilor Shono Yukihira. Sharp-tongued workaholic with keen observational skills. Complicit in framing colleague Yamazaki Kai for bribery which led to his suicide. Knows many secrets of the councilor and worries her involvement and the concealment of the truth would lead to retaliation.\\

\textbf{Kikuo Inagaki} \\
The station master of Ofumoto Station, a steady and experienced manager who is diligent and knowledgeable about the station's operations and takes good care of passengers. Real identity is Yamazaki Kocho, father of framed victim Yamazaki Kai. Threatening station attendant Masako to cooperate with him while plotting vengeance. \\

\textbf{Masako Ueno} \\
Young station attendant coerced by imposter station master. Attempts to secretly warn Kogoro under the threat of the fake station master through cryptic messages while maintaining nervous demeanor, which makes her a bit distracted at work. \\

\midrule

\multicolumn{1}{c}{\textbf{Scene 1 - Seven People in the Waiting Room}} \\
\midrule
\textbf{Background:} A sudden typhoon disrupts the travel plans of several individuals at Ofumoto Station, causing all train services to be suspended indefinitely. The waiting room is filled with the passengers and station staff, whose paths are about to intertwine in unexpected ways.\\
\textbf{Location:} Ofumoto Station Waiting Room \\
\textbf{Characters:}\\
\quad - Conan\\
\quad - Kogoro Mouri: Wants to inquire about train schedule.\\
\quad - Yuichi: Attempting loan sales pitch.\\
\quad - Morris: Proposing to enjoy the hot spring in the station and attempting to steal from others.\\
\quad - Hitoshi: Lamenting life failures, hates salesmen.\\
\quad - Noriko: Being angry and aggressive due to the disrupted work schedule. \\
\quad - Inagaki: Monitoring Masako while maintaining professional demeanor.\\
\quad - Masako: Being watched by the impostor station master, so acting nervous.\\
\textbf{Plots:}\\
\quad - Kogoro Mouri remarks on the sudden typhoon, and the stationmaster and employee distribute towels to everyone.\\
\quad - Tension erupts between Hitoshi and Yuichi over loan scams.\\
\quad - Yuichi's briefcase accident reveals multiple company brochures.\\
\quad - Noriko reveals political connections through an outburst.\\
\quad - Discovery of severed communication lines forces everyone to stay overnight.\\
\quad - Decision to stay overnight leads to a hot spring proposal.\\
\midrule
\multicolumn{1}{c}{\textbf{Scene 2 - Dangerous Signal}} \\
\midrule
\textbf{Background:} The passengers, who were forced to spend the night at the station due to the disrupted train schedules, seek temporary relief in the station's hot spring facility. However, some individuals have ulterior motives and plan to take advantage of the situation.\\
\textbf{Location:} Ofumoto Station Hot Spring \\
\textbf{Characters:}\\
\quad - Conan\\
\quad - Kogoro Mouri: Deducing the identity of the person who wrote the note.\\
\quad - Yuichi: Leaving early as he is worried that others might find the advertisements in his bag at the locker.\\
\quad - Morris: Leaving early as he wants to steal from the lockers while others being in the hot spring.\\
\textbf{Plots:}\\
\quad - After the storm, the few people in the hot springs enjoy a brief moment of comfort.\\
\quad - Morris and Yuichi leave first. Yuichi mentions that he's worried about the items in his locker because it doesn't have a lock.\\
\quad - Kogoro Mouri finds a note in his locker, which reads: ``Mr. Mouri, someone will be killed here tonight. The murderer is...''.\\
\quad - Kogoro Mouri and Conan discuss who might have written the note and realize that everyone is a suspect.\\
\quad - Kogoro Mouri decides not to tell anyone about the note for the time being and to find out who wrote it first.\\
\midrule
\multicolumn{1}{c}{\textbf{Scene 3 - Limited Investigation}} \\
\midrule
\textbf{Background:} After discovering an anonymous note hinting a murder tonight, Kogoro Mouri and Conan decide to conduct a secret investigation by having one-on-one talk with others to uncover the truth behind the ominous warning.\\
\textbf{Location:} Ofumoto Station Waiting Room \\
\textbf{Characters:}\\
\quad - Conan\\
\quad - Ran Mouri: You gaze out at the raging typhoon outside the waiting room and can't help but reminisce about the trip to Osaka with Shinichi two years ago, when you were also stranded at a station due to a typhoon. You spent an unforgettable time together. But now, Shinichi is missing, and you don't know when you'll be able to travel with him again, which makes you very sad.\\
\quad - Yuichi: After coming out of the hot springs, you specifically checked the luggage in your locker. You are very worried that someone might find the various company brochures in your luggage and realize that you are a fraud. You have a hunch that Morris is interested in your luggage, so you are particularly wary of him. Therefore, when Conan asks you questions, you inadvertently imply that Morris is suspicious and seems to have some secrets, in order to divert suspicion from yourself.\\
\quad - Morris: The real reason you suggested that everyone go to the hot springs was to find an opportunity to steal valuable items from their luggage. However, you heard that Kogoro Mouri is a famous detective and dared not touch his or Conan's belongings. You left the hot springs early to check Yuichi's luggage, but to your surprise, Yuichi came out with you. You are worried that Yuichi might have realized your true identity.\\
\quad - Hitoshi: The thought of ending your life has been lingering in your mind. Therefore, while everyone was in the hot springs, you sat in the waiting room without interacting with anyone.\\
\quad - Noriko: While everyone was in the hot springs, you sat in the waiting room eating snacks and trying to contact the councilor. Unfortunately, there is almost no signal in the station, and Hitoshi is also in the waiting room.\\
\quad - Masako:  You previously wrote a note to Mr. Mouri, warning him that someone would be killed here tonight, but you didn't have time to write the station master's name. Now you are serving the passengers in the waiting room, and you clearly notice that the fake station master is watching you from not far away in the waiting hall. Therefore, you are very nervous and dare not make any sudden moves. You really want to tell Mr. Mouri or Conan that you wrote the note, but you don't have the courage.\\
\quad - Inagaki: You continue to play the role of an amiable and responsible station master. At the same time, you are happy to share with others about your son, Akihiro, who has a stable job and family. Of course, all of this is an act. You want to show others that you are a family-loving father to avoid arousing suspicion about your true identity. Only Masako in the waiting room knows that you are an imposter, and you are constantly monitoring her behavior. To the outside world, you show concern for her as a senior. At the same time, you want to find out what Masako told Conan, but you know that this will arouse suspicion, so you are very cautious. Once you know that the news has leaked, you are very clear that it must be from Masako, but you still pretend to be very surprised.\\
\textbf{Plots:}\\
\quad - Yuichi:\\
\quad\quad - Hints at his skepticism of Maurice.\\
\quad - Morris:\\
\quad\quad - When Conan asks you, subtly imply that Yuichi is suspicious.\\
\quad\quad - Share your travel experiences.\\
\quad - Hitoshi:\\
\quad\quad - Lament your life.\\
\quad - Noriko:\\
\quad\quad - Proof that Hitoshi stays in the waiting room, mumbling about his life and failures.\\
\quad\quad - Based on the conversations and behavior of everyone in the waiting room earlier, you deduce that Yuichi is a scammer and Morris is a pickpocket.\\
\quad - Masako:\\
\quad\quad - Provides the service he deserves with enthusiasm.\\
\quad - Ran Mouri:\\
\quad\quad - Reminisces with Conan about his past with Shinichi.\\
\quad\quad - Comforts Conan.\\

\bottomrule
\caption{Drama script for the demonstration.}
\end{longtable}

\section{Aggressive Agent Player}
\label{player_agents}
\begin{longtable}{p{3cm} p{12.4cm}}
\toprule

Grumpy Guy      & You were reluctantly dragged into this Detective Conan interactive drama. You’re not particularly interested, and your interactions throughout the game reflect your impatience and mild irritation. \\ \midrule

Fan Girl        & You're a passionate Detective Conan fan and a second-year university student. You're obsessed with the charming male characters in the series and are incredibly excited to play this interactive drama. During the game, your curiosity gets the best of you, and you often ask in-game characters about off-topic subjects, like Heiji Hattori, Shinichi Kudo, Tooru Amuro, and Kaito Kid. \\ \midrule

Confused Man    & You're a high school student who was dragged into this Detective Conan interactive drama without knowing anything about the series. You feel completely out of place and confused, often relying on in-game characters for guidance as you try to figure out what’s going on. \\ \midrule

Strolling Lady  & You have an unusual quirk. Despite knowing that it’s snowing heavily outside in the game, you still look for every opportunity to invite in-game characters to take a walk with you. \\ \midrule

Flamer          & You’re fascinated by how characters react to conflicts. Throughout the game, you carefully observe conversations between characters, constantly looking for opportunities to stir up tension or fan the flames of existing disputes. \\ \midrule

Screenwriter    & You're a professional writer specializing in mystery and detective fiction, and you joined this Detective Conan interactive drama in search of creative inspiration. You’re highly interested in the game’s narrative structure, character motivations, and emotional depth. Throughout the game, you meticulously analyze every detail of the story. \\ \midrule

Heartbroken One & You’re a university student who recently went through a breakup. You joined this Detective Conan interactive drama hoping for some emotional distraction or even comfort from the characters. While you do follow the mystery plot, you occasionally steer conversations toward romantic subplots, to subtly reflect on your own experiences. \\ \midrule

Troublemaker    & You're a mischievous middle school student who joined this Detective Conan interactive drama out of sheer curiosity. While you’re familiar with the series, you intentionally make nonsensical or low-intelligence choices during the game just to see how others react and stir up some chaos. \\ \midrule

Multilingual    & You’re a foreign language learner who enjoys showing off your linguistic skills. During the game, you frequently throw in sentences in different languages, such as Spanish or Japanese, to demonstrate your fluency. \\ \midrule

Demanding       & Throughout the game, you are extremely unhappy about the train service being suspended due to the snow. You insist that the station master and staff find a way to get you back to Tokyo, leading to frequent conflicts. \\ \bottomrule
\caption{Personalities of aggressive player agents.}
\end{longtable}

\end{document}